\documentclass[11pt,a4paper]{article}
\usepackage[hyperref]{acl2021}
\usepackage{times}
\usepackage[autostyle, english = american]{csquotes}
\MakeOuterQuote{"}
\usepackage{latexsym}

\usepackage{tabularx}

\usepackage{microtype}
\usepackage{graphicx}
\usepackage{caption}
\aclfinalcopy % Uncomment this line for the final submission
\title{Towards the evaluation of automatic simultaneous speech translation from a communicative perspective}

\author{Claudio Fantinuoli \\
  Mainz University/KUDO \\
  \texttt{fantinuoli@uni-mainz.de} \\\And
  Bianca Prandi \\
  Mainz University \\
  \texttt{prandi@uni-mainz.de} \\}

\date{}

\begin{document}
\maketitle
\begin{abstract}
In recent years, automatic speech-to-speech and speech-to-text translation has gained momentum thanks to advances in artificial intelligence, especially in the domains of speech recognition and machine translation. The quality of such applications is commonly tested with automatic metrics, such as BLEU, primarily with the goal of assessing improvements of releases or in the context of evaluation campaigns. However, little is known about how the output of such systems is perceived by end users or how they compare to human performances in similar communicative tasks. 

In this paper, we present the results of an experiment aimed at evaluating the quality of a real-time speech translation engine by comparing it to the performance of professional simultaneous interpreters. To do so, we adopt a framework developed for the assessment of human interpreters and use it to perform a manual evaluation on both human and machine performances. In our sample, we found better performance for the human interpreters in terms of intelligibility, while the machine performs slightly better in terms of informativeness. The limitations of the study and the possible enhancements of the chosen framework are discussed. Despite its intrinsic limitations, the use of this framework represents a first step towards a user-centric and communication-oriented methodology for evaluating real-time automatic speech translation.
\end{abstract}

\section{Introduction} \label{Introduction}

Real-time or simultaneous speech translation (ST) aims at translating a continuous speech input from one language to another with the lowest latency\footnote{In this context, we broadly define latency as the time delay from when an utterance is pronounced in the source language to when it gets translated in the target language.} and highest quality possible. In recent years, automatic speech translation systems have been developed at scale, and their quality has improved significantly \citep{sperber_speech_2020}. At present, research is increasingly focusing on end-to-end trainable encoder-decoder models, i.e. speech-to-speech (STS) or speech-to-text (STT) translation systems that directly match source and target language \citep{di_gangi_fine-tuning_2018, jia_direct_2019, ansari_findings_2020}. Nonetheless, the cascading approach is de facto still the mainstream solution for speech translation (ST). The main reason is that this approach benefits from the remarkable improvements in automatic speech recognition (ASR) \citep{chiu_state---art_2018} and machine translation (MT) \citep{barrault_findings_2020} obtained thanks to the wealth of task-speciﬁc data available. In cascading systems, the process of translating from speech to text or from speech to speech is performed by a series of concatenated modules. In most cases, these systems apply ASR to the speech input, and then pass the results on to an MT engine. Since a short latency is an important characteristic of such systems, the translation is rendered while the source is unfolding, on the basis of different approaches ranging from simple time delay to complex agents that establish when the context is sufficient to perform the translation. Several additional components can be integrated into this pipeline, such as text normalization \citep{fugen_system_2008}, suppression of speech disfluencies \citep{fitzgerald_reconstructing_2009}, prosody transfer \citep{kano_end--end_2018}, and so forth.

Real-time ST systems have the potential to be used in communicative settings, such as institutional events, lectures, conferences, etc. in order to make multilingual content accessible in real-time, thus increasing inclusion and participation when human services for language accessibility are not available, such as live interlingual subtitling \citep{romero-fresco_quality_2017} or conference  interpreting \citep{pochhacker_introducing_2016}. So far, the evaluation of ST in general, and real-time ST in particular, has been framed in the domain of computer science (CS). In CS, automatic metrics are applied in order to compare systems and monitor progress over time\footnote{See for example the methodology used for the shared task of the International Conference on Spoken Language Translation 2021 available at \url{https://iwslt.org/2021/simultaneous}}. However, little is known about how such systems, that for the sake of this paper we will define as machine interpreting (MI) systems\footnote{We tentatively define as Machine Interpreting all automatic methods of real-time speech translation, i.e. cascading, end-to-end, into text, into speech, etc. that are used in the context of real-life communication.}, perform in real communication settings and whether they are able to meet the needs of end users. To the best of our knowledge, no evaluation framework has been developed and deployed in the past to assess the performance of such systems from a communicative perspective.

To address this shortcoming, in the present contribution we apply a user-centric evaluation framework derived from Interpreting Studies (IS) to the task of assessing an automatic system for ST. Moving towards an evaluation framework that takes into consideration the authentic communicative setting, we compare the performances of the automated system with the performances of professional simultaneous interpreters. We do so in order to assess the level of usability of such framework and to benchmark the performances of the machine, inferring the suitability of the ST system for the proposed communication task from its comparison with the human performance.

The rest of the paper is organised as follows. In Section \ref{RelatedWork}, we present an overview of research areas in the field of automatic speech translation and human interpreting evaluation. In Section \ref{Methodology}, we illustrate our research methodology and the experimental design. Section \ref{Dataset} describes the dataset created for this task, while Section \ref{Framework} introduces the framework used to evaluate the performance of the machine and of the human interpreters. Section \ref{Results} presents the results of the evaluation and Section \ref{Discussion} discusses and puts the results into perspective. Section \ref{Conclusion} concludes the paper with final remarks.

\section{Related work} \label{RelatedWork}

The evaluation of simultaneous speech translation, independently of whether the process is performed by a human or a machine, is a topic central both to the domain of Computer Science and of Interpreting Studies.

In CS, ST is typically evaluated in terms of quality and latency. Similar to MT, the approach used consists in the application of automatic metrics in order to allow for a fast and objective evaluation of the systems \citep{ma_simuleval_2020}. However, due to its novelty, the ST research community currently lacks a universally adopted evaluation methodology. Quality is generally measured by BLEU \citep{papineni_bleu:_2002,post_call_2018}, TER \citep{snover_study_2006} and METEOR \citep{banerjee_meteor_2005}. The approach to compare system outputs against source texts, gold standard translations, and other system outputs represents, despite the limitations of such metrics \citep{babych_automated_2014}, a widely accepted evaluation methodology. The measurement of latency, which broadly corresponds to the ear-voice span of human interpreting \citep[e.g.][]{gile_basic_2009}, represents a more challenging task that still lacks sufficient clarity and consistency. In this context, several metrics have been introduced, such as Average Proportion (AP) \citep{cho_can_2016}, Continues Wait Length (CW) \citep{gu_learning_2017}, Average Lagging (AL) \citep{ma_simuleval_2020}, Differentiable Average Lagging (DAL) \citep{cherry_thinking_2019}. Generally speaking, the evaluation approach used in CS is product-oriented. The concept of quality is limited to measuring proximity in the linguistic surface between translation and ground truth. It does not take into consideration the user perception, the pragmatic aspect of communication, and, intrinsically, cannot consider the translation process as embedded in a communicative event \citep[e.g.][]{angelelli_interpretation_2002}. 

This is different to IS. Since human interpretation always occurs in a specific communicative setting, the need to evaluate it accordingly has always been in focus. Here, the pursuit of conceptual and methodological tools for the empirical study and assessment of quality has a long tradition, particularly in the conference domain and simultaneous modality \citep[e.g.][]{pochhacker_quality_2002, kalina_quality_2005, mikkelson_quality_2015}. Despite the different perspectives that have been adopted to define and evaluate quality, there is considerable agreement among scholars on a number of criteria which are considered fundamental when evaluating human interpretation. Most criteria of quality are associated with the product-oriented perspective and can be subsumed in two main areas, the first one focusing primarily on the interpretation or target-text as "a `faithful' image" \citep{gile_basic_2009} or "exact and faithful reproduction" \citep{jones_conference_2002} of the original speech, the second one on the notion of intelligibility, also called clarity, target-text comprehensibility, linguistic acceptability, stylistic correctness, etc. Such evaluation is centred on the view of interpreting as a language processing task. At an even higher level, quality can also be seen under the paradigm of a holistic idea of successful communication. From this perspective, interpreting is assessed on the basis of whether it successfully allows the parties involved in a particular context of interaction to achieve their communicative goal, as judged from the various perspectives in and on the communicative event \citep{gile_basic_2009}. The focus of this perspective is no longer on the product (the rendition), but rather on the communicative action performed to achieve a certain purpose and effect, and therefore on the holistic function of facilitating communicative interaction \citep{pochhacker_quality_2002}. 

From a methodological perspective, quality in interpretation has been evaluated through surveys \citep{feldweg_konferenzdolmetscher_1996}, measures of performance through experimentation \citep{shlesinger_stranger_1995}, or corpus-based analysis \citep{russo_corpus-based_2018}. Different to the CS approach, which is based on automatic metrics, the analysis of data in IS is performed on the basis of a manual evaluation of the corpus data. 

While such evaluation frameworks have been designed and used regularly in the domain of machine translation, very few attempts have been made so far to evaluate the performance of automatic speech translation system both in the context of the product-based and of the holistic/communicative approach. A few pilot studies on the usability of ST systems have only been performed in the context of dialogue interpreting \citep{curten_maschinelles_2016, wonisch_skype_2017}, while only one has been attempted in the area of real-time ST \citep{muller_evaluation_2016}. We believe that such approaches, if appropriately adapted to the research desideratum at hand, could contribute to a better understanding and evaluation of machine speech translation systems.

\section{Data and methodology} \label{Methodology}
As discussed in the previous section, ST systems are typically evaluated by means of automatic metrics using reference datasets. Although such evaluations are useful to compare systems among each other, one of their main limitations is that they do not take into consideration the communicative setting nor the perception of their usefulness by final users. To overcome this limitation, we select and apply to the assessment of ST a user-centric framework commonly used for the evaluation of human interpretation. 

In order to understand the potential usefulness of the automatically generated translation, we compare the machine performance with a gold standard: the interpretation delivered by professional human interpreters in the real context of the event. Simultaneous interpretation (SI) is the modality most commonly used to provide multilingual access in real-time\footnote{The other would be interlingual respeaking for the creation of live subtitling which is, however, still in its infancy.}. Since we assume that the service provided by professional interpreters allows communication among the parties in the event, we consider it to be our "communicative" ground truth. This gold standard is not an ideal rendition of the original, but it comes with all the benefits and limitations of the real simultaneous translation used at a specific event to overcome language barriers. By means of this comparison we can infer, at least to some extent, the communicative performance of the machine in the context of a real communicative event. The overall question driving our research is therefore "How does the performance of a speech-to-text translation system compare with human SI?". 

To answer this question we compile a corpus of speeches in English delivered in real-life contexts and align them with their human interpreted versions into Italian as well as with the output of a simultaneous STT translation system chosen for this task. The dataset is described in Section \ref{Dataset}. We manually assess the quality of the human and automatic renditions (transcriptions) on the basis of the evaluation framework described in Section \ref{Framework}. This evaluation represents an attempt to apply a more user-centric approach to the assessment of the automatic service provided by STT translation systems.

\subsection{Dataset} \label{Dataset}

There are several speech translation corpora currently available, such as MuST-C \citep{di_gangi_must-c_2019} and Europarl-ST \citep{iranzo-sanchez_europarl-st_2020}. They generally contain source speeches in one language and the corresponding written or, in a few cases, spoken translations in the target language(s). While they are useful to explore end-to-end ST, for example to train the language models, they have not been designed with the goal of assessing such systems from a communicative perspective. As a matter of fact, the target language component of the corpus is in most cases an edited translation, therefore a product of mediated, offline, and decontextualized work. 

To overcome this limitation, we create a new pilot corpus of speeches (lectures) and of their live translations which would allow us to conduct a better evaluation of the machine output by comparing it with the gold standard produced by humans in a real communicative event (see Section \ref{Methodology}).

The main rationale behind the creation of our corpus is the selection of naturally occurring data on which to conduct our observation, both for the original speech and, most importantly, for the gold standard (the basis of the comparison). 
%Rather than defining a priori a set of criteria for the selection of speeches to include in the corpus, we pick a random set of events on which to run the STT engine, and apply our baseline evaluation. 

The five speeches selected for the corpus are randomly extracted from two series of talks ("Festival dell'economia" and "Meeting di Rimini") that had been originally interpreted simultaneously from English into Italian by five different interpreters. Both the original speeches and their interpretations are publicly available on the web\footnote{ \url{https://www.festivaleconomia.it/} and \url{https://www.meetingrimini.org}}. After choosing the events, 2-minute extracts are randomly selected from each speech. The small size of the corpus does not allow for generalizations, but should provide indications on the suitability of the chosen evaluation framework. With this approach, the ecological validity is maximal, as the research data are drawn from real interpreted events, while the level of control is minimal, making data harder to interpret, especially when it comes to causality \citep{baekelandt_elicitation_2020}.

The speeches included in the corpus\footnote{The corpus is available at  \url{https://cai.uni-mainz.de/steval}.} are summarised in Table \ref{tab:corpus}. While all the speakers had presented in English, three were native speakers (texts 1, 3 and 5) and two were not (texts 2 and 4). As for the source text delivery mode, four speakers presented "impromptu" speeches (texts 1, 2 , 4 and 5) and one a "read-aloud" speech (text 3). The topics included: economy (text 1), bit coin (text 2), artificial intelligence (text 3), green growth (text 4) and medicine (text 5), with different degrees of technicality. The audio quality was good for all speeches. The speed of delivery ranges from 142 to 160 words per minute (wpm) and is in line with typical speech rates at conferences \citep[e.g.][]{ehrensberger-dow_cognitive_2015}. 

Similar to \citeauthor{batista_recovering_2008} (\citeyear{batista_recovering_2008}), the corpus for evaluation is presented in written form. Since the output of the STT is already produced by the engine as written text, only the source speeches and the human interpretations are transcribed by means of an ASR engine and manually corrected. The ST output is included in the corpus without modifications. The five texts are segmented in utterances and aligned with the interpretations. The number of segments for each text ranges from 16 to 20.

\begin{table}[h]
\centering
 \begin{tabular}{||c c c c||} 
 \hline
 Text & Duration & Words & Speed (wpm) \\ [0.5ex] 
 \hline\hline
 1 & 2' 10'' & 347 & 160\\ 
 \hline
 2 & 2' 02'' & 288 & 142\\
 \hline
 3 & 2' 00'' & 320 & 160\\
 \hline
 4 & 2' 01'' & 304 & 157\\
 \hline
 5 & 2' 07'' & 320 & 151\\ [1ex] 
 \hline
\end{tabular}
\caption{\label{tab:corpus} Corpus features }
\end{table}   

For this experiment, we choose the real-time ST service offered by Azure Speech Translation \footnote{\url{https://azure.microsoft.com/en-en/services/cognitive-services/speech-translation/}}. The main reason for this choice is that the service is available as a commercial API and represents the state-of-the art of cascading systems. Different to human interpreters, who deliver the translation orally, this API translates speech into written text without any form of speech synthesis. In principle, this generates an asymmetry in the evaluation. However, since the selected framework requires the evaluation to be performed on the written transcriptions, this lack of symmetry has been deemed as non central for the purpose of this experiment.

To collect the data of the ST engine, a simple Web application was created by the authors around the API. The application sends the original speech to the API and records the real-time translation returned by the service. Because the evaluation is performed on written transcriptions, in this experiment the latency of the system was not taken into account, and only the final translation hypothesis generated was considered for the evaluation. This is a major limitation of this study that needs to be addressed in future experiments.

\subsection{Evaluation framework and procedure} \label{Framework}
For the investigation and the comparison of the human and the machine output, an evaluation framework derived from the Interpreting Studies \citep{angelelli_revisiting_2009} is chosen and slightly adapted. The framework is "assumed to account for central aspects of the interpreted event but not for its entirety as a communicative event" \citep[p. 99]{angelelli_revisiting_2009}. As discussed in Section \ref{Introduction}, at this stage we follow a product-based approach to quality assessment in IS, leaving the situated evaluation of the interpreted event for later explorations, for which an extended framework comprising additional communicative perspectives and criteria should be defined. Notwithstanding the limitations of this approach, one of the advantages of Tiselius's framework against automatic metrics lies in its being user-centric and in line with the corpus-based evaluation already established in Interpreting Studies to assess the quality of human interpretation. 

Tiselius defines the framework as "an easy-to-use tool that can be implemented by laypeople in order to assess a transcribed version of a simultaneous interpreting performance" \citep[p. 99]{angelelli_revisiting_2009}. This aspect is particularly important for possible future use of the framework. In order to further streamline it,  we slightly simplified the evaluation scale, and adapted its wording in order to make it suitable to express a judgement on both human and automatic speech translation. 

The framework aims at assessing the target production on the basis of two dimensions:

\begin{itemize}
    \item Intelligibility, defined as the evaluation of the target text in terms of fluency, clarity, adequacy etc., performed without a comparison with the source text
    \item Informativeness, defined as the evaluation of the target text in terms of semantic information content, performed with a comparison with the source text
\end{itemize}

The two dimensions reflect the main criteria at the core of the product-oriented approach to quality evaluation in IS (Section \ref{RelatedWork}). 6 raters with a background in interpreting and translation are asked to conduct the evaluation of the human interpretation (HI) and the machine output (MI). For each speech, the raters are asked to assess on a six-point Likert scale first the intelligibility of the HI and of the MI output (without a comparison with the source speech nor a comparison between the two outputs), then to evaluate the informativeness of the two renditions (HI and MI) by comparing each one to the source speech.

While this methodology represents a first step towards a more holistic approach to the evaluation of ST, it also presents a series of shortcomings:

\begin{itemize}
    \item The product-based evaluation of the gold standard, the HI, is conducted on transcriptions and not on the audio output. Not only do prosody, modulation of voice, hesitations, etc. constitute distinctive aspects of spoken (human) language, but they are also actively used by human interpreters to reach several communicative goals. They contribute, for example, to disambiguate oral speech, explicate references, etc. The evaluation on the basis of transcriptions deprives the evaluator of these key features, with obvious negative implications for the quality scores. A viable option could be to perform the evaluation on the basis of an audio corpus, thus retaining all the features of spoken language during the evaluation of the human interpreters. Another promising way to address this shortcoming would be to resort to interlingual respeaking as a gold standard instead of HI. Since the output of respeakers is a written rendition of the original in the target language, it would make the output of human and automatic ST more comparable. 
    \item Notwithstanding the efforts made to keep the framework as simple as possible, the evaluation procedure proves quite time-consuming. Conducting evaluation campaigns on a bigger scale with this framework may be hampered by this aspect.   
    \item The item definitions in the six-point scale are not sufficiently straightforward to guide the rater in taking a decision. Further simplification and rewording are required.
    \item The assessment does not take into consideration latency, which is important to judge the real-time translation at a communicative level, especially as far as the user experience is concerned. The ST system used in the experiment, for example, performed real-time adaptations on the target language, i.e. modifying the translation hypothesis while receiving increasing context from the source speech. The impact on comprehension and user friendliness of both this aspect and disfluencies in the human rendition should be studied more attentively in future.
    \item As will become clear in Section \ref{Results}, resorting to human interpretation instead of written translation as a gold standard calls for new strategies in the evaluation. Because human interpreters and ST systems perform the task using a different approach (linear for machines, interpretive for humans), the comparison using classical methodologies may be inadequate.  

\end{itemize}    
The shortcomings of the evaluation framework should be addressed in a follow-up study.

\section{Results} \label{Results}
The scores for each of the two parameters (intelligibility and informativeness) are summed for each speech, output and rater and then averaged. The relative percentage score is calculated on the maximum amount of points obtainable for each text. The figures below illustrate the results of data analysis.

\begin{figure}[t]
\includegraphics[width=8cm]{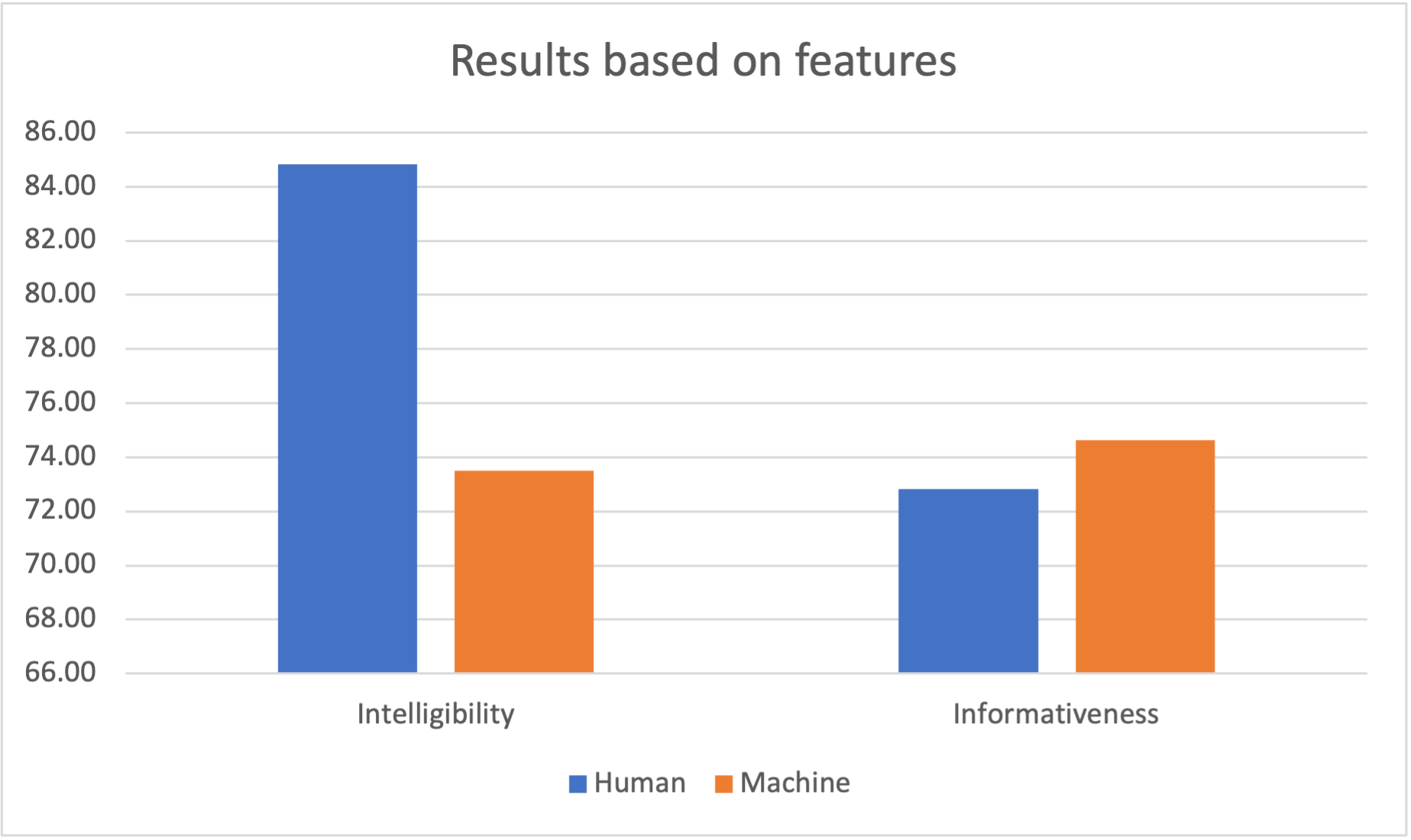}
\centering
\caption{Intelligibility and informativeness scores for the human and the machine output.}
\label{fig:results_features}
\end{figure}

As shown by Figure \ref{fig:results_features}, human interpreters obtain better scores for intelligibility (84.84 \% to 73.49 \%), while the machine output is rated slightly better than the human interpretations in terms of informativeness (74.63 \% to 72.82 \%).

\begin{figure}[t]
\includegraphics[width=8cm]{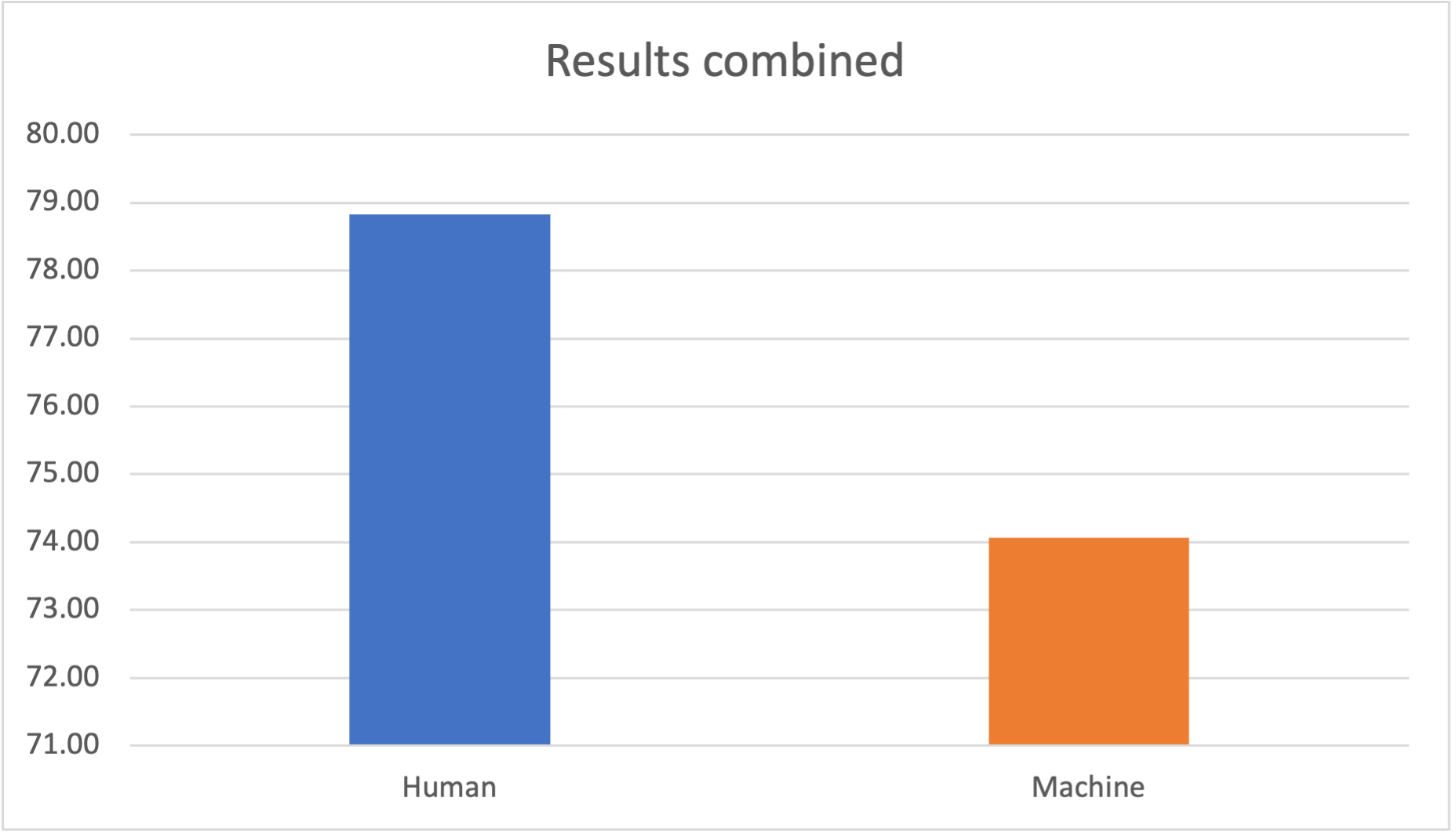}
\centering
\caption{Scores combined for the human and the machine output.}
\label{fig:results_combined}
\end{figure}

When combining the scores for the two rating criteria (Figure \ref{fig:results_combined}), the human output surpasses machine output by 4.77 percentage points. It can be argued that the two parameters do not have the same weight in terms of their impact on the success of the communicative event. At this stage of evaluation, however, we decide to combine them without any weight and to leave this more in-depth analysis to a later phase of development of our evaluation framework.

\begin{figure}[t]
\includegraphics[width=8cm]{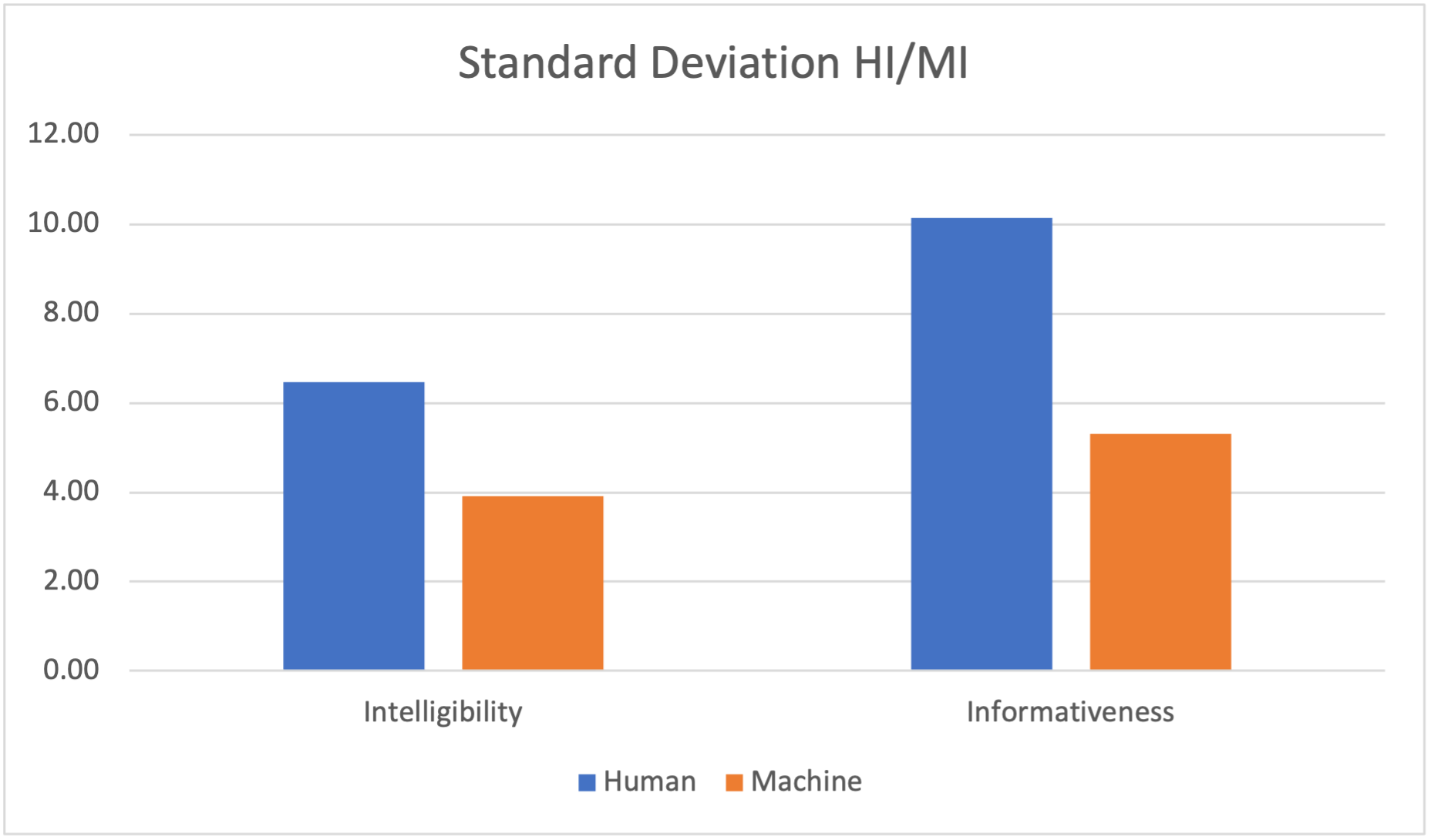}
\centering
\caption{Standard deviation of the intelligibility and informativeness scores (human and machine output).}
\label{fig:SD}
\end{figure}

Figure \ref{fig:SD} illustrates the standard deviation (SD) for the two evaluation parameters for all 5 speeches. Both for the criterion intelligibility and for the criterion informativeness, the SD for the human output is larger than for the machine output, for which the scores are very close to each other in our sample.

This result suggests that variables such as topic, density, speed of the original speech, accents, etc. affect less the machine than the human interpreter. On the one hand, this is quite surprising if one considers that aspects such as performance of ASR with foreign accents, to name but one example, are considered detrimental in automatic language processing \citep{kitashov_foreign_2018, shi_accented_2021}. On the other hand, the larger SD for the human output may point to the fact that humans tend to have a high degree of variance in performances due to different background knowledge, skills, etc. Because of the small size of the corpus, the trends observed in the present study cannot be generalized and the analysis should be conducted on a larger sample.

In order to verify the adequacy of the evaluation methodology, and in particular of the rating scales used to assess intelligibility and informativeness achieved by the human and the machine interpretation, Krippendorff's $\alpha$ is calculated for the two evaluation criteria and the two types of output. This measure is chosen as it allows to overcome the problems presented by Fleiss's $\kappa$ \citep{hayes_answering_2007}, another common measure of intercoder reliability used when multiple raters are involved \citep[see][]{mellinger_quantitative_2016}. The statistic is interpreted like other measures of reliability, with higher scores indicating higher intercoder reliability. Overall, $\alpha$ values are below the lowest value (.667) defined as acceptable by \citet{krippendorff_content_2013} for tentative conclusions, and well below the recommended value of .800 (ibid.). The $\alpha$ is considerably lower than these values for both intelligibility and informativeness in human interpreting ($\alpha$ = .442 and .607 respectively). The $\alpha$ for MI intelligibility is .658, while the value of inter-rater reliability for the category of informativeness in MI is barely acceptable ($\alpha$ = .676). Overall, these results suggest that applying the evaluation scale derived from IS \textit{as is} for the comparison of HI and MI output presents limitations that need to be addressed in future work, for instance in terms of the optimisation of the scoring rubric and the inclusion of further dimensions in the evaluation scale.

\section{Discussion} \label{Discussion}
The differences in the raters' evaluation of the human and machine output can be better understood by analysing several phenomena retrieved from the corpus. The complexity of the evaluation and, inherently, of the comparison between human and machine interpretation is strictly linked with the pragmatic nature of HI, which often calls for interventions on the part of the interpreter. Such interventions, emerging on the linguistic surface of the interpreted text, may be evidence of underlying strategic behaviour exercised, for instance, to favour comprehension, or may be the result of emergency coping tactics aimed at preventing a disruption of the rendition in adverse conditions, for instance in the case of particularly information-dense or fast speeches. Phenomena such as generalisation, addition and (intentional) omission \citep[see for instance][]{gile_basic_2009, kohn_strategic_2002} seem to occur more often in human SI than in written translation, and are entirely absent in the automatic translation of speech. The MT engine lacks any linguistic phenomena that may index intentional interventions, not only because it lacks deliberateness, but also because it has been trained on written (and not interpreted) texts. This fundamental difference may limit the ability of a classic evaluation framework (both manual and automatic) to provide an assessment of quality which reflects the communicative success of an event mediated by human or by machine interpretation. In order to illustrate our argument, we report several example passages from the corpus complete of their rendition by the human interpreters and by the ST engine.

In the following example (Table \ref{tab:addition}), the human interpreter added a reference to the financial crisis ("momento della crisi finanziaria") implicit in the temporal reference provided by the speaker (2009). At the same time, one unit of information ("where a lot of people were looking for this phrase") was left out by the human interpreter, while it is present in the machine output.

\begin{table}[h]%
\caption{\label{tab:addition} Addition }
\begin{tabularx}{\linewidth}{ c X }
S & {\emph{So you have this spike around 2009 where a lot of people were looking for this phrase}}\\
\hline
HI & {perché nel 2009 abbiamo un picco, momento della crisi finanziaria}\\
\hline
MI & {quindi avete questo picco intorno al 2009 o molte persone stavano cercando questa frase}\\
\end{tabularx}
\end{table}%

In the following example (Table \ref{tab:generalisation}), the interpreter opted for a generalisation: "we've not spent enough energy, time, and money" was summed up in "we really have to do more", which conveys the same key message while making explicit what is meant by the original speaker, but is less precise than the automatic rendition, more adherent to the source text. 

\begin{table}[h]%
\caption{Generalisation\label{tab:generalisation}}
\begin{tabularx}{\linewidth}{ c X }
S & {\textit{It's clear that we've not spent enough energy, time, and money in protecting our healtcare workers}}\\\hline
HI & {Quindi ecco. Dobbiamo veramente fare di più per proteggere i nostri operatori sanitari}\\\hline
MI & {È chiaro che non abbiamo speso abbastanza energia, tempo e denaro per proteggere i nostri operatori sanitari}\\\hline\end{tabularx}
\end{table}%

The two examples discussed above illustrate an inherent conundrum in the evaluation, i.e. how to evaluate pragmatic interventions by the interpreter. Whether such interventions should be considered as justifiable or not and which rendition, the human or the machine, is more appreciated by the end-user and more conducive to the same communicative goal pursued by the speaker cannot be reflected in an evaluation framework such as the one chosen for this initial exploration. Furthermore, these phenomena substantiate our argument that a framework for the evaluation of
ST in comparison with HI requires a broader perspective.

Another key point of comparison between HI and ST lies in the presence and evaluation of errors. It may be argued that blatant errors are more apparent in ST than in professional HI, as exemplified by the following passage:

\begin{table}[h]%
\caption{\label{tab:blatant}Blatant error in MI}
\begin{tabularx}{\linewidth}{ c X }
S & {\textit{So 9 billion of dollars have been raised through ICOs}}\\\hline
HI & {per cui ci sono adesso 9 miliardi di dollari che sono stati raccolti attraverso queste operazioni di ICO}\\\hline
MI & {Quindi 9 miliardi di dollari sono stati raccolti attraverso i devoti ghiacciati}\\\hline
\end{tabularx}
\end{table}%

The erroneous translation in the automatic output ("devoti ghiacciati", i.e. iced pious), clearly due to a speech recognition issue (Example \ref{tab:blatant}), is immediately recognisable as such by the (human) end-user. This type of blatant mistake seems to be a distinctive characteristic of MI and is more frequent than in neural machine translation because of the key features of oral speech. It would be interesting to explore the effects of this error type in a real-life communicative event. However, human interpreters may also commit severe mistakes. Let us consider the following case:

\begin{table}[h]%
\caption{\label{tab:blatant2}Blatant error in HI}
\begin{tabularx}{\linewidth}{ c X }
S & {\textit{we have lots of historical examples of overestimating how fast it will kick in}}\\\hline
HI & {Ci sono esempi storici in questo senso di sottovalutazione della velocità in cui le cose sono cambiate}\\\hline
MI & {abbiamo molti esempi storici di sopravvalutazione della velocità con cui prenderà il via}\\\hline
\end{tabularx}
\end{table}%

At first sight, the HI may appear more elegant and fluent than the automatic output (the median intelligibility score for this segment is 5). Thus, the wrong rendition of "overestimating" with "sottovalutazione" (EN: underestimating), due to erroneous anticipation or to having misheard the speaker's words, may go unnoticed without a comparison with the source text.

The examples discussed above emphasise on the one hand that the type of mistakes end-users are confronted with may be of very different nature. The effects of the various types of mistakes on communication and their evaluation by human raters may also vary, and should be explored within a framework that takes into account the communicative perspective. On the other hand, this comparison also stresses the need to compare ST not with the ideal of HI but with the variability of human performances.

\section{Conclusion and Future Work} \label{Conclusion}
This paper reports on an experiment that compares the output of a real-time speech-to-text translation system with the performance of human interpreters. The main goal was to expand the methodology that is used nowadays to evaluate such systems from the purely computational approach based on automatic metrics to a more user-centric and communication-oriented one. To do so, we apply an evaluation framework derived from Interpreting Studies and let six evaluators assess the performance of humans and machines according to the criteria of intelligibility and informativeness. The results show a better performance by humans in terms of intelligibility and a slightly better performance by the machine in terms of accuracy. 

Despite several drawbacks of the framework adopted, the path initiated with this study may bear fruits in terms of better understanding and evaluating the output of speech-to-text and speech-to-speech translation systems in the context of situated multilingual communication and its pragmatic context. The study also highlights several limitations of the approach chosen. They are mainly related to the difficulty of defining objective criteria in the evaluation of quality of interpreted texts, and to the intrinsic shortcomings of evaluating a communicative event only on the basis of the product of the translation process without the contextual embedding of the evaluation in the communicative setting. Such shortcomings need to be addressed in future work.

\bibliographystyle{acl_natbib}
\bibliography{anthology,acl2021}

\begin{thebibliography}{44}
\expandafter\ifx\csname natexlab\endcsname\relax\def\natexlab#1{#1}\fi

\bibitem[{Angelelli(2002)}]{angelelli_interpretation_2002}
Claudia Angelelli. 2002.
\newblock \href {https://doi.org/10.7202/001891ar} {Interpretation as a
  {Communicative} {Event}: {A} {Look} through {Hymes}' {Lenses}}.
\newblock \emph{Meta}, 45(4):580--592.

\bibitem[{Ansari et~al.(2020)Ansari, Axelrod, Bach, Bojar, Cattoni, Dalvi,
  Durrani, Federico, Federmann, Gu, Huang, Knight, Ma, Nagesh, Negri, Niehues,
  Pino, Salesky, Shi, Stüker, Turchi, Waibel, and Wang}]{ansari_findings_2020}
Ebrahim Ansari, Amittai Axelrod, Nguyen Bach, Ondřej Bojar, Roldano Cattoni,
  Fahim Dalvi, Nadir Durrani, Marcello Federico, Christian Federmann, Jiatao
  Gu, Fei Huang, Kevin Knight, Xutai Ma, Ajay Nagesh, Matteo Negri, Jan
  Niehues, Juan Pino, Elizabeth Salesky, Xing Shi, Sebastian Stüker, Marco
  Turchi, Alexander Waibel, and Changhan Wang. 2020.
\newblock \href {https://doi.org/10.18653/v1/2020.iwslt-1.1} {{FINDINGS} {OF}
  {THE} {IWSLT} 2020 {EVALUATION} {CAMPAIGN}}.
\newblock In \emph{Proceedings of the 17th {International} {Conference} on
  {Spoken} {Language} {Translation}}, pages 1--34, Online. Association for
  Computational Linguistics.

\bibitem[{Babych(2014)}]{babych_automated_2014}
Bogdan Babych. 2014.
\newblock \href {https://doi.org/10.5565/rev/tradumatica.70} {Automated {MT}
  evaluation metrics and their limitations}.
\newblock \emph{Tradumàtica: tecnologies de la traducció}, (12):464.

\bibitem[{Baekelandt and Defrancq(2020)}]{baekelandt_elicitation_2020}
Annelies Baekelandt and Bart Defrancq. 2020.
\newblock \href {https://doi.org/10.1080/0907676X.2020.1849322} {Elicitation of
  particular grammatical structures in speeches for interpreting research:
  enhancing ecological validity of experimental research in interpreting}.
\newblock \emph{Perspectives}, pages 1--18.

\bibitem[{Banerjee and Lavie(2005)}]{banerjee_meteor_2005}
Satanjeev Banerjee and Alon Lavie. 2005.
\newblock \href {https://www.aclweb.org/anthology/W05-0909} {{METEOR}: {An}
  {Automatic} {Metric} for {MT} {Evaluation} with {Improved} {Correlation} with
  {Human} {Judgments}}.
\newblock In \emph{Proceedings of the {ACL} {Workshop} on {Intrinsic} and
  {Extrinsic} {Evaluation} {Measures} for {Machine} {Translation} and/or
  {Summarization}}, pages 65--72, Ann Arbor, Michigan. Association for
  Computational Linguistics.

\bibitem[{Barrault et~al.(2020)Barrault, Biesialska, Bojar, Costa-jussà,
  Federmann, Graham, Grundkiewicz, Haddow, Huck, Joanis, Kocmi, Koehn, Lo,
  Ljubešić, Monz, Morishita, Nagata, Nakazawa, Pal, Post, and
  Zampieri}]{barrault_findings_2020}
Loïc Barrault, Magdalena Biesialska, Ondřej Bojar, Marta~R. Costa-jussà,
  Christian Federmann, Yvette Graham, Roman Grundkiewicz, Barry Haddow,
  Matthias Huck, Eric Joanis, Tom Kocmi, Philipp Koehn, Chi-kiu Lo, Nikola
  Ljubešić, Christof Monz, Makoto Morishita, Masaaki Nagata, Toshiaki
  Nakazawa, Santanu Pal, Matt Post, and Marcos Zampieri. 2020.
\newblock \href {https://www.aclweb.org/anthology/2020.wmt-1.1} {Findings of
  the 2020 {Conference} on {Machine} {Translation} ({WMT20})}.
\newblock In \emph{Proceedings of the {Fifth} {Conference} on {Machine}
  {Translation}}, pages 1--55, Online. Association for Computational
  Linguistics.

\bibitem[{Batista et~al.(2008)Batista, Caseiro, Mamede, and
  Trancoso}]{batista_recovering_2008}
Fernando Batista, Diamantino Caseiro, Nuno Mamede, and Isabel Trancoso. 2008.
\newblock \href {https://doi.org/10.1016/j.specom.2008.05.008} {Recovering
  capitalization and punctuation marks for automatic speech recognition: {Case}
  study for {Portuguese} broadcast news}.
\newblock \emph{Speech Communication}, 50(10):847--862.

\bibitem[{Bendazzoli(2018)}]{russo_corpus-based_2018}
Claudio Bendazzoli. 2018.
\newblock \href {https://doi.org/10.1007/978-981-10-6199-8_1} {Corpus-based
  {Interpreting} {Studies}: {Past}, {Present} and {Future} {Developments} of a
  ({Wired}) {Cottage} {Industry}}.
\newblock In Mariachiara Russo, Claudio Bendazzoli, and Bart Defrancq, editors,
  \emph{Making {Way} in {Corpus}-based {Interpreting} {Studies}}, pages 1--19.
  Springer Singapore, Singapore.
\newblock Series Title: New Frontiers in Translation Studies.

\bibitem[{Cherry and Foster(2019)}]{cherry_thinking_2019}
Colin Cherry and George Foster. 2019.
\newblock \href {http://arxiv.org/abs/1906.00048} {Thinking {Slow} about
  {Latency} {Evaluation} for {Simultaneous} {Machine} {Translation}}.
\newblock \emph{arXiv:1906.00048 [cs]}.
\newblock ArXiv: 1906.00048.

\bibitem[{Chiu et~al.(2018)Chiu, Sainath, Wu, Prabhavalkar, Nguyen, Chen,
  Kannan, Weiss, Rao, Gonina, Jaitly, Li, Chorowski, and
  Bacchiani}]{chiu_state---art_2018}
Chung-Cheng Chiu, Tara~N. Sainath, Yonghui Wu, Rohit Prabhavalkar, Patrick
  Nguyen, Zhifeng Chen, Anjuli Kannan, Ron~J. Weiss, Kanishka Rao, Ekaterina
  Gonina, Navdeep Jaitly, Bo~Li, Jan Chorowski, and Michiel Bacchiani. 2018.
\newblock \href {http://arxiv.org/abs/1712.01769} {State-of-the-art {Speech}
  {Recognition} {With} {Sequence}-to-{Sequence} {Models}}.
\newblock \emph{arXiv:1712.01769 [cs, eess, stat]}.
\newblock ArXiv: 1712.01769.

\bibitem[{Cho and Esipova(2016)}]{cho_can_2016}
Kyunghyun Cho and Masha Esipova. 2016.
\newblock \href {http://arxiv.org/abs/1606.02012} {Can neural machine
  translation do simultaneous translation?}
\newblock \emph{arXiv:1606.02012 [cs]}.
\newblock ArXiv: 1606.02012.

\bibitem[{Collados~Aís and García~Becerra(2015)}]{mikkelson_quality_2015}
Ángela Collados~Aís and Olalla García~Becerra. 2015.
\newblock Quality.
\newblock In Holly Mikkelson and Renee Jourdenais, editors, \emph{The
  {Routledge} handbook of interpreting}, Routledge {Handbooks} in {Applied}
  {Linguistics}. Routledge, London ; New York.

\bibitem[{Cürten(2016)}]{curten_maschinelles_2016}
Giulia Cürten. 2016.
\newblock \emph{Maschinelles {Dolmetschen} mit {Google} Übersetzer.}
\newblock Ph.D. thesis, University of Vienna.

\bibitem[{Di~Gangi et~al.(2019)Di~Gangi, Cattoni, Bentivogli, Negri, and
  Turchi}]{di_gangi_must-c_2019}
Mattia~A. Di~Gangi, Roldano Cattoni, Luisa Bentivogli, Matteo Negri, and Marco
  Turchi. 2019.
\newblock \href {https://doi.org/10.18653/v1/N19-1202} {{MuST}-{C}: a
  {Multilingual} {Speech} {Translation} {Corpus}}.
\newblock In \emph{Proceedings of the 2019 {Conference} of the {North}}, pages
  2012--2017, Minneapolis, Minnesota. Association for Computational
  Linguistics.

\bibitem[{Di~Gangi et~al.(2018)Di~Gangi, Dessì, Cattoni, Negri, and
  Turchi}]{di_gangi_fine-tuning_2018}
Mattia~Antonino Di~Gangi, Roberto Dessì, Roldano Cattoni, Matteo Negri, and
  Marco Turchi. 2018.
\newblock \href {http://arxiv.org/abs/1810.07652} {Fine-tuning on {Clean}
  {Data} for {End}-to-{End} {Speech} {Translation}: {FBK} @ {IWSLT} 2018}.
\newblock \emph{arXiv:1810.07652 [cs, eess, stat]}.
\newblock ArXiv: 1810.07652.

\bibitem[{Feldweg(1996)}]{feldweg_konferenzdolmetscher_1996}
Erich Feldweg. 1996.
\newblock \emph{Der {Konferenzdolmetscher} im internationalen
  {Kommunikationsprozeß}}.
\newblock Julius Groos, Heidelberg.
\newblock Bibtex: feldweg\_konferenzdolmetscher\_1996.

\bibitem[{Fitzgerald et~al.(2009)Fitzgerald, Hall, and
  Jelinek}]{fitzgerald_reconstructing_2009}
Erin Fitzgerald, Keith Hall, and Frederik Jelinek. 2009.
\newblock Reconstructing {False} {Start} {Errors} in {Spontaneous} {Speech}
  {Text}.
\newblock In \emph{Proceedings of the 12th {Conference} of the {European}
  {Chapter} of the {ACL} ({EACL} 2009)}, pages 255--263. Association for
  Computational Linguistics.

\bibitem[{Fügen(2008)}]{fugen_system_2008}
Christian Fügen. 2008.
\newblock \emph{A {System} for {Simultaneous} {Translation} of {Lectures} and
  {Speeches}.}
\newblock Ph.D. thesis, University of Karlsruhe.

\bibitem[{Gile(2009)}]{gile_basic_2009}
Daniel Gile. 2009.
\newblock \href {http://www.jbe-platform.com/content/books/9789027288080}
  {\emph{Basic {Concepts} and {Models} for {Interpreter} and {Translator}
  {Training}: {Revised} edition}}, 2nd edition.
\newblock John Benjamins Publishing Company, Amsterdam.

\bibitem[{Gu et~al.(2017)Gu, Neubig, Cho, and Li}]{gu_learning_2017}
Jiatao Gu, Graham Neubig, Kyunghyun Cho, and Victor O.~K. Li. 2017.
\newblock \href {http://arxiv.org/abs/1610.00388} {Learning to {Translate} in
  {Real}-time with {Neural} {Machine} {Translation}}.
\newblock \emph{arXiv:1610.00388 [cs]}.
\newblock ArXiv: 1610.00388.

\bibitem[{Hayes and Krippendorff(2007)}]{hayes_answering_2007}
Andrew~F. Hayes and Klaus Krippendorff. 2007.
\newblock \href {https://doi.org/10.1080/19312450709336664} {Answering the
  {Call} for a {Standard} {Reliability} {Measure} for {Coding} {Data}}.
\newblock \emph{Communication Methods and Measures}, 1(1):77--89.

\bibitem[{Iranzo-Sánchez et~al.(2020)Iranzo-Sánchez, Silvestre-Cerdà, Jorge,
  Roselló, Giménez, Sanchis, Civera, and
  Juan}]{iranzo-sanchez_europarl-st_2020}
Javier Iranzo-Sánchez, Joan~Albert Silvestre-Cerdà, Javier Jorge, Nahuel
  Roselló, Adrià Giménez, Albert Sanchis, Jorge Civera, and Alfons Juan.
  2020.
\newblock \href {http://arxiv.org/abs/1911.03167} {Europarl-{ST}: {A}
  {Multilingual} {Corpus} {For} {Speech} {Translation} {Of} {Parliamentary}
  {Debates}}.
\newblock \emph{arXiv:1911.03167 [cs, eess]}.
\newblock ArXiv: 1911.03167.

\bibitem[{Jia et~al.(2019)Jia, Weiss, Biadsy, Macherey, Johnson, Chen, and
  Wu}]{jia_direct_2019}
Ye~Jia, Ron~J. Weiss, Fadi Biadsy, Wolfgang Macherey, Melvin Johnson, Zhifeng
  Chen, and Yonghui Wu. 2019.
\newblock \href {http://arxiv.org/abs/1904.06037} {Direct speech-to-speech
  translation with a sequence-to-sequence model}.
\newblock \emph{arXiv:1904.06037 [cs, eess]}.
\newblock ArXiv: 1904.06037.

\bibitem[{Jones(2002)}]{jones_conference_2002}
Roderick Jones. 2002.
\newblock \emph{Conference {Interpreting} {Explained}}.
\newblock Routledge, Manchaster.

\bibitem[{Kalina(2005)}]{kalina_quality_2005}
Sylvia Kalina. 2005.
\newblock \href {https://doi.org/10.7202/011017ar} {Quality {Assurance} for
  {Interpreting} {Processes}}.
\newblock \emph{Meta: Journal des traducteurs}, 50(2):768.

\bibitem[{Kano et~al.(2018)Kano, Takamichi, Sakti, Neubig, Toda, and
  Nakamura}]{kano_end--end_2018}
Takatomo Kano, Shinnosuke Takamichi, Sakriani Sakti, Graham Neubig, Tomoki
  Toda, and Satoshi Nakamura. 2018.
\newblock \href {https://doi.org/10.1007/s10590-018-9217-7} {An end-to-end
  model for cross-lingual transformation of paralinguistic information}.
\newblock \emph{Machine Translation}, 32(4):353--368.

\bibitem[{Kitashov et~al.(2018)Kitashov, Svitanko, and
  Dutta}]{kitashov_foreign_2018}
Fedor Kitashov, Elizaveta Svitanko, and Debojyoti Dutta. 2018.
\newblock \href {http://arxiv.org/abs/1807.03625} {Foreign {English} {Accent}
  {Adjustment} by {Learning} {Phonetic} {Patterns}}.
\newblock \emph{arXiv:1807.03625 [cs, eess, stat]}.
\newblock ArXiv: 1807.03625.

\bibitem[{Kohn and Kalina(2002)}]{kohn_strategic_2002}
Kurt Kohn and Sylvia Kalina. 2002.
\newblock \href {https://doi.org/10.7202/003333ar} {The {Strategic} {Dimension}
  of {Interpreting}}.
\newblock \emph{Meta}, 41(1):118--138.

\bibitem[{Krippendorff(2013)}]{krippendorff_content_2013}
Klaus Krippendorff. 2013.
\newblock \emph{Content analysis: an introduction to its methodology}, 3rd ed
  edition.
\newblock SAGE, Los Angeles ; London.

\bibitem[{Ma et~al.(2020)Ma, Dousti, Wang, Gu, and Pino}]{ma_simuleval_2020}
Xutai Ma, Mohammad~Javad Dousti, Changhan Wang, Jiatao Gu, and Juan Pino. 2020.
\newblock \href {http://arxiv.org/abs/2007.16193} {{SimulEval}: {An}
  {Evaluation} {Toolkit} for {Simultaneous} {Translation}}.
\newblock \emph{arXiv:2007.16193 [cs]}.
\newblock ArXiv: 2007.16193.

\bibitem[{Mellinger and Hanson(2016)}]{mellinger_quantitative_2016}
Christopher~D. Mellinger and Thomas Hanson. 2016.
\newblock \href {https://doi.org/10.4324/9781315647845} {\emph{Quantitative
  {Research} {Methods} in {Translation} and {Interpreting} {Studies}}}.
\newblock Routledge, London.

\bibitem[{Müller et~al.(2016)Müller, Fünfer, Stüker, and
  Waibel}]{muller_evaluation_2016}
Markus Müller, Sarah Fünfer, Sebastian Stüker, and Alex Waibel. 2016.
\newblock \href
  {http://www.lrec-conf.org/proceedings/lrec2016/summaries/677.html}
  {Evaluation of the {KIT} {Lecture} {Translation} {System}.}
\newblock In \emph{Proceedings of the {Tenth} {International} {Conference} on
  {Language} {Resources} and {Evaluation} {LREC} 2016, {Portorož}, {Slovenia},
  {May} 23-28, 2016.}

\bibitem[{Papineni et~al.(2002)Papineni, Roukos, Ward, and
  Zhu}]{papineni_bleu:_2002}
Kishore Papineni, Salim Roukos, Todd Ward, and Wei-Jing Zhu. 2002.
\newblock {BLEU}: {A} {Method} for {Automatic} {Evaluation} of {Machine}
  {Translation}.
\newblock In \emph{Proceedings of the 40th {Annual} {Meeting} of the
  {Association} for {Computational} {Linguistics} ({ACL})}, pages 311--318.

\bibitem[{Post(2018)}]{post_call_2018}
Matt Post. 2018.
\newblock \href {https://doi.org/10.18653/v1/W18-6319} {A {Call} for {Clarity}
  in {Reporting} {BLEU} {Scores}}.
\newblock In \emph{Proceedings of the {Third} {Conference} on {Machine}
  {Translation}: {Research} {Papers}}, pages 186--191, Belgium, Brussels.
  Association for Computational Linguistics.

\bibitem[{Pöchhacker(2002)}]{pochhacker_quality_2002}
Franz Pöchhacker. 2002.
\newblock \href {https://doi.org/10.7202/003847ar} {Quality {Assessment} in
  {Conference} and {Community} {Interpreting}}.
\newblock \emph{Meta}, 46(2):410--425.

\bibitem[{Pöchhacker(2016)}]{pochhacker_introducing_2016}
Franz Pöchhacker. 2016.
\newblock \emph{Introducing {Interpreting} {Studies}}, 2nd edition.
\newblock Routledge.

\bibitem[{Romero-Fresco and Pöchhacker(2017)}]{romero-fresco_quality_2017}
Pablo Romero-Fresco and Franz Pöchhacker. 2017.
\newblock \href
  {https://lans-tts.uantwerpen.be/index.php/LANS-TTS/article/view/438} {Quality
  assessment in interlingual live subtitling: {The} {NTR} {Model}}.
\newblock \emph{Linguistica Antverpiensia, New Series – Themes in Translation
  Studies}, 16.

\bibitem[{Seeber(2015)}]{ehrensberger-dow_cognitive_2015}
Kilian~G. Seeber. 2015.
\newblock \href {https://benjamins.com/catalog/bct.72.03see} {Cognitive load in
  simultaneous interpreting: {Measures} and methods}.
\newblock In Maureen Ehrensberger-Dow, Susanne Göpferich, and Sharon O'Brien,
  editors, \emph{Benjamins {Current} {Topics}}, volume~72, pages 18--33. John
  Benjamins Publishing Company, Amsterdam.
\newblock Bibtex: seeber\_cognitive\_2015.

\bibitem[{Shi et~al.(2021)Shi, Yu, Lu, Liang, Feng, Wang, Qian, and
  Xie}]{shi_accented_2021}
Xian Shi, Fan Yu, Yizhou Lu, Yuhao Liang, Qiangze Feng, Daliang Wang, Yanmin
  Qian, and Lei Xie. 2021.
\newblock \href {http://arxiv.org/abs/2102.10233} {The {Accented} {English}
  {Speech} {Recognition} {Challenge} 2020: {Open} {Datasets}, {Tracks},
  {Baselines}, {Results} and {Methods}}.
\newblock \emph{arXiv:2102.10233 [cs, eess]}.
\newblock ArXiv: 2102.10233.

\bibitem[{Shlesinger(1995)}]{shlesinger_stranger_1995}
Miriam Shlesinger. 1995.
\newblock \href {https://doi.org/10.1075/target.7.1.03shl} {Stranger in
  {Paradigms}: {What} {Lies} {Ahead} for {Simultaneous} {Interpreting}
  {Research}?}
\newblock \emph{Target}, 7(1):7--28.
\newblock Bibtex: shlesinger\_stranger\_1995.

\bibitem[{Snover et~al.(2006)Snover, Dorr, Schwartz, Micciulla, and
  Makhoul}]{snover_study_2006}
Matthew Snover, Bonnie Dorr, Richard Schwartz, Linnea Micciulla, and John
  Makhoul. 2006.
\newblock A study of translation edit rate with targeted human annotation.
\newblock In \emph{In {Proceedings} of {Association} for {Machine}
  {Translation} in the {Americas}}, pages 223--231.

\bibitem[{Sperber and Paulik(2020)}]{sperber_speech_2020}
Matthias Sperber and Matthias Paulik. 2020.
\newblock \href {http://arxiv.org/abs/2004.06358} {Speech {Translation} and the
  {End}-to-{End} {Promise}: {Taking} {Stock} of {Where} {We} {Are}}.
\newblock \emph{arXiv:2004.06358 [cs]}.
\newblock ArXiv: 2004.06358.

\bibitem[{Tiselius(2009)}]{angelelli_revisiting_2009}
Elisabet Tiselius. 2009.
\newblock \href {https://doi.org/10.1075/ata.xiv.07tis} {Revisiting {Carroll}'s
  scales}.
\newblock In Claudia~V. Angelelli and Holly~E. Jacobson, editors,
  \emph{American {Translators} {Association} {Scholarly} {Monograph} {Series}},
  volume XIV, pages 95--121. John Benjamins Publishing Company, Amsterdam.

\bibitem[{Wonisch(2017)}]{wonisch_skype_2017}
Alexander Wonisch. 2017.
\newblock \emph{Skype {Translator}: {Funktionsweise} und {Analyse} der
  {Dolmetschleistung} in der {Sprachrichtung} {Englisch}-{Deutsch}.}
\newblock Ph.D. thesis, Wonisch, Alexander (2017) Skype Translator:
  Funktionsweise und Analyse der Dolmetschleistung in der Sprachrichtung
  Englisch-Deutsch. Masterarbeit, University of Vienna.

\end{thebibliography}

%\appendix

\end{document}